\documentclass{article}
\usepackage{iclr2027_conference,times}
\iclrfinalcopy
\makeatletter
\renewcommand{\footnotesize}{\@setfontsize\footnotesize\@viiipt{9.5}}
\makeatother

\usepackage{booktabs}
\usepackage{graphicx}
\usepackage{amsmath,amssymb,amsfonts,bm}
\usepackage{placeins}
\usepackage{float}
\usepackage{multirow}
\usepackage{array}
\usepackage{tabularx}
\usepackage{enumitem}
\usepackage{capt-of}
\usepackage{xcolor}
\usepackage{colortbl}
\usepackage{tikz}
\usepackage{needspace}
\usepackage{indentfirst}
\usepackage{cuted}
\usepackage{currfile}
\usepackage{caption}
\usepackage{balance}
\usepackage{stfloats}
\usetikzlibrary{arrows.meta,positioning,shapes.geometric,fit,backgrounds}

\newcolumntype{Y}{>{\raggedright\arraybackslash}X}
\newcolumntype{C}{>{\centering\arraybackslash}X}
\newcolumntype{H}{>{\hsize=\dimexpr 3\hsize+4\tabcolsep\relax\centering\arraybackslash}X}

\definecolor{tableblue}{RGB}{226,233,254}
\definecolor{tablelightblue}{RGB}{227,243,254}
\definecolor{tablegray}{RGB}{242,242,242}
\definecolor{tabgroupa}{RGB}{233,244,240}
\definecolor{tabgroupb}{RGB}{230,237,246}
\definecolor{deltagood}{RGB}{0,140,70}
\definecolor{deltabad}{RGB}{200,50,50}
\newcommand{\mhDeltaImproveUp}[1]{\smash{\textcolor{deltagood}{\ensuremath{\scriptstyle\uparrow\!#1}}}}
\newcommand{\mhDeltaImproveDown}[1]{\smash{\textcolor{deltagood}{\ensuremath{\scriptstyle\downarrow\!#1}}}}

\newif\ifmhCompareDeltas\mhCompareDeltastrue
\makeatletter
\let\mh@origrowcolor\rowcolor
\renewcommand{\rowcolor}[1]{\mh@origrowcolor{#1}[\dimexpr\tabcolsep+0.8pt\relax][\dimexpr\tabcolsep+0.8pt\relax]}
\makeatother

\AtBeginDocument{%
  \setlength{\abovedisplayskip}{-1pt plus0pt minus0pt}%
  \setlength{\belowdisplayskip}{-1pt plus0pt minus0pt}%
  \setlength{\abovedisplayshortskip}{-1pt plus0pt minus0pt}%
  \setlength{\belowdisplayshortskip}{-1pt plus0pt minus0pt}%
}

\graphicspath{{./}{figures/}}
\usepackage[breaklinks,colorlinks]{hyperref}
\newcommand{\SuppSecappaboveprocapability}{\SuppSecappclaimscopeparked}

\newcommand{\SuppSecappadditionalresults}{Supp.\ Section~1.6}
\newcommand{\SuppSecsecsuppmixbudgetladder}{\SuppSecappadditionalresults}

\newcommand{\SuppSecappalternateprovidertransfer}{\SuppSecappadditionalresults}

\newcommand{\SuppSecappcasestudies}{Supp.\ Section~1.7}
\newcommand{\SuppSecappclaimscopeparked}{Supp.\ Section~1.8}
\newcommand{\SuppSecappcoinedarmprotocols}{Supp.\ Section~1.3}
\newcommand{\SuppSecappcomponentcontracts}{Supp.\ Section~1.5}
\newcommand{\SuppSecappdecisioncalculus}{Supp.\ Section~1.4}
\newcommand{\SuppSecappreproducibilityboundary}{Supp.\ Section~1.9}
\newcommand{\SuppSecappruntimegatechecklist}{\SuppSecappdecisioncalculus}

\newcommand{\SuppSecappruntimeobjects}{Supp.\ Section~1.1}
\newcommand{\SuppSecappscrubbedrouting}{Supp.\ Section~1.2}
\newcommand{\SuppSecappuactivationpolicysamepool}{\SuppSecappadditionalresults}

\newcommand{\SuppSecappwhylivenemix}{\SuppSecappadditionalresults}

\title{MedHarness: A Unified Budgeted Runtime for Reliable Medical Imaging Agents}

\author{Haofan Wu \\
University of Birmingham
\And
Le Zhang \\
University of Birmingham
}

\begin{document}
\maketitle
\fancyhead{}
\fancyfoot[C]{\thepage}
\renewcommand{\headrulewidth}{0pt}
\setlength{\headwidth}{\textwidth}

\begin{abstract}
Medical imaging agents already return answers on heterogeneous imaging surfaces, yet deployment still faces a limited verification budget where each wrong accept can harm clinical trust. When a thin check and a deeper evidence pass disagree on the same case, deciding when to stop, open more evidence, repair, or reject without a shared controller remains ill-posed. Tool-using workflows often lock a surface-specific evidence order per dataset, or a private always-open full-depth verification path; those pipelines stay costly, leave many wrong accepts, and cannot share one controller across modalities. We present \textbf{MedHarness}, a registered multi-surface runtime reliability layer that reformulates post-answer control as shared decision-object attachment after a proposer's initial prediction. Under Shared Decision Object, Minimum-Sufficient Verification, and Intermediate Repair, one shared policy with surface recipe adapters lets a new imaging surface join the same stop / open-evidence / repair / reject controller by recipe registration. On a 512-case live pool, MedHarness cuts wrong accepts versus a full-verification baseline by about 47\% and raises accepted accuracy to 83.2\% vs.\ 72.9\% at 86.1\% coverage versus full release under that baseline. On a failure-enriched multi-surface stress ranking, the same controller cuts estimated remote service calls per case by about 71\% versus that baseline under a matched context-load setting. A held-out three-surface check and an alternate language-model transfer keep the same control ordering. The code is available at \url{https://anonymous.4open.science/r/MedHarness-paper-release-A7F3/}.
\end{abstract}

\section{Introduction}
\label{sec:intro}
Medical vision-language models and medical agents increasingly produce plausible initial predictions for image-grounded tasks \cite{mmedagent,llavamed}. At deployment, those answers still need a runtime control process that decides when to stop and accept, when to open more evidence, when to repair, and when to reject \cite{geifman2017selective,jiang2021trust}. Each wrong accept can harm clinical trust, while checking every case at full depth is rarely affordable under a limited verification budget. Without that shared control process, a strong first answer can still be released unsafely, and a weak answer can still be over-checked at constant cost. Clinical-impact studies likewise emphasize that predictive accuracy alone does not determine whether an individual output is safe to accept \cite{mckinney2020breast,kelly2019clinical}. When a thin check and a deeper evidence pass disagree on the same case, deciding when to stop and accept, when to open more evidence, when to repair, and when to reject without one shared controller remains ill-posed.

Retrieval-augmented and tool-using medical workflows improve grounding \cite{medagents,wu2025medagentpro}. They still often lock a preset evidence path that is a surface-specific tool order per dataset, or a private always-open full-depth verification path per dataset, when the imaging surface changes \cite{lewis2020rag,yao2023react}. Selective prediction and self-refine modules exist \cite{chow1970reject,madaan2023selfrefine}, but are rarely composed as one registered, budgeted runtime across heterogeneous medical imaging queues---so each dataset tends to grow its own pipeline and control decisions stay unshared. This raises a central systems question: after a proposer returns an initial prediction, how should a single controller decide stop, open-evidence, repair, and reject under one limited budget across registered surfaces? We reformulate that question as attaching each case to a shared decision object that carries the candidate, evidence state, and terminal action under surface recipe adapters.

We present \textbf{MedHarness}, a registered multi-surface runtime reliability layer that wraps a static proposer with those adapters and a shared control policy. Fig.~\ref{fig:runtime_overview} summarizes the shared controller and surface recipe attachment. Under three complementary principles---Shared Decision Object, Minimum-Sufficient Verification, and Intermediate Repair---a new imaging surface joins the same stop / open-evidence / repair / reject controller by recipe registration, without rewriting a private always-open full-depth verification path per dataset. The contribution is the budgeted runtime control process; it starts from the proposer's initial prediction and treats verification, repair, and abstention as control actions over that shared decision object.

Our contributions: (i)~We propose \textbf{MedHarness}, a registered multi-surface runtime reliability layer that, under Shared Decision Object, Minimum-Sufficient Verification, and Intermediate Repair, attaches surface recipe adapters to one shared stop / open-evidence / repair / reject controller so heterogeneous imaging surfaces join by recipe registration; (ii)~on a 512-case live pool, MedHarness cuts wrong accepts versus a full-verification baseline by about 47\% through abstention and raises accuracy among accepted answers to 83.2\% vs.\ 72.9\% at 86.1\% coverage (71 abstentions) versus full release under that baseline, and on a failure-enriched four-surface stress ranking cuts estimated remote service calls per case by about 71\% and estimated context-load token counts by about 69\% versus that baseline, with live ablations isolating repair, specialist evidence, and lock; (iii)~a held-out three-surface check and an alternate language-model transfer keep MedHarness above full verification and above the proposer, and the budgeted controller improves wrong-accept rate and accuracy among accepted answers over full-verification and thin-check baselines, while the unit module charge falls versus full verification on the live pool (3.86 versus 4.00) and on the four-surface stress ranking (3.15 versus 4.00), separate from the remote service-call cut of request events; the thin-check charge is 1.99, so this module-charge cut stays separate from that unit charge on~this~pool.

\section{Related Work}
\label{sec:related}

% End Introduction-local ICLR \parskip; body keeps preamble \parskip.
\setlength{\parskip}{0pt}

\textbf{Imaging agents.} Medical imaging agents and task-specific visual assessors have long been built from segmentation stacks to self-configuring biomedical pipelines~\cite{unet,nnunet}. High-stakes skin and retinal settings further motivate careful release of image-grounded decisions~\cite{esteva2017dermatologist,gulshan2016dr}. Medical foundation models and multimodal assistants raise first-pass answer quality on imaging questions~\cite{medsam,llavamed}. Vision--language pretraining and instruction tuning push toward reusable multi-task systems~\cite{radford2021clip,liu2023llava}. Clinical-language and generalist medical AI studies add pressure toward general medical reasoning~\cite{singhal2023clinical,moor2023generalist}. These lines improve candidate generation, yet leave open how one shared runtime should stop, open evidence, repair, or reject after an already-formed imaging answer under a limited verification budget. MedHarness attaches that shared controller after a static proposer via surface recipe adapters.

\textbf{Selective prediction.} Selective prediction and reject options learn when to abstain so wrong accepts do not enter the scored set~\cite{elyaniv2010selective,geifman2017selective}. Classical reject bounds further motivate abstention when the answerer is unreliable~\cite{chow1970reject}. Trustworthy machine learning surveys likewise treat when-to-stop as a first-class systems concern~\cite{jiang2021trust,hendrycks2017baseline}. These mechanisms supply refuse pieces, but are rarely composed with open-evidence and intermediate repair as one registered controller across heterogeneous imaging queues. MedHarness treats stop, open-evidence, repair, and reject as typed actions over one shared decision object under Shared Decision Object, Minimum-Sufficient Verification, \mbox{and Intermediate Repair together.}

\textbf{Budgeted tools and repair.} Retrieval-augmented and tool-learning agents expose external evidence or computation~\cite{lewis2020rag,schick2023toolformer}. Action-oriented planners and reflective multi-agent frameworks make decomposition and collaboration operational~\cite{yao2023react,reflexion}. Self-refine and self-consistency add correction loops on the answer itself~\cite{madaan2023selfrefine,wang2023selfconsistency}. Medical agentic systems bring collaboration and multimodal tool use into clinical-style workflows~\cite{medagents,kim2024mdagents}. Imaging agents that use tools and evidence-linked planners still often lock in advance a private workflow or an always-open full-depth verification path per surface~\cite{wu2025medagentpro,fallahpour2025medrax}. Process-level verification catches stepwise errors that outcome scores miss~\cite{lightman2024verify,gou2024critic}. These systems supply modules and traces a budgeted controller can call; MedHarness composes them under one registered stop / open-evidence / repair / reject interface so a new imaging surface can join the same stop, open-evidence, repair, and reject loop.

\section{Method}
\label{sec:method}

% Page-top float placement: queued immediately so it opens the next page;
% Method prose below fills under the figure. Natural full-width only.
\begin{figure*}[t]
\setlength{\floatsep}{4pt plus0pt minus1pt}%
\setlength{\textfloatsep}{6pt plus0pt minus2pt}%
\setlength{\intextsep}{6pt plus0pt minus2pt}%
\centering
\includegraphics[width=\textwidth,keepaspectratio]{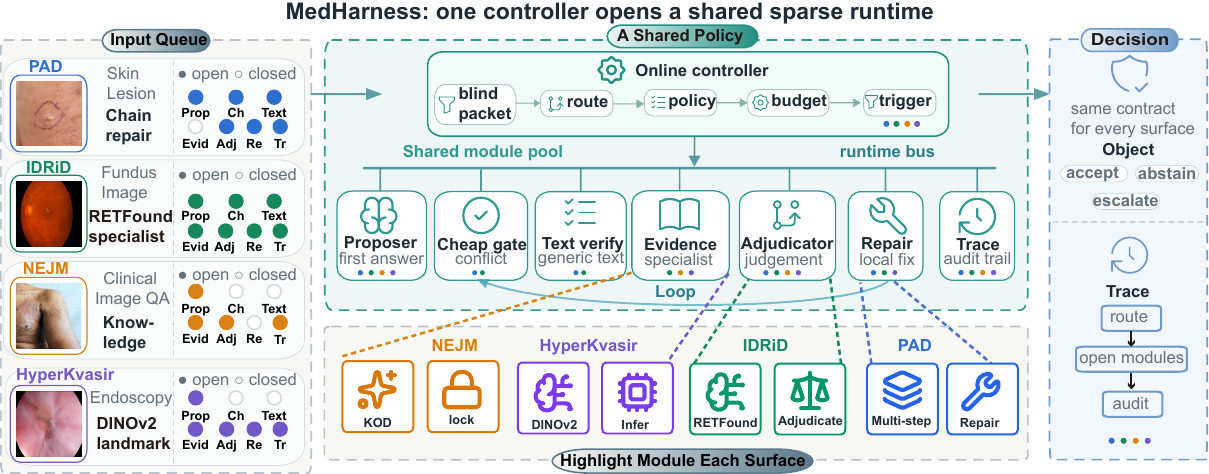}
\caption{Shared runtime from clinical input through budgeted controller and shared module pool to one terminal decision object. Colored dots mark which surfaces use each card; ink stubs evidence (Evid), check (Ch), adjust (Adj), transfer (Tr), and knowledge on demand (KOD); open or stop detail \SuppSecappcoinedarmprotocols{}.}
\label{fig:runtime_overview}
\end{figure*}

We aim to attach a budgeted runtime reliability layer after a static proposer's
initial prediction, so medical imaging agents decide when to stop, when to open
failure-mode evidence, when to repair, and when to abstain or escalate under one
shared registration map across heterogeneous surfaces~\cite{camel}.
Fig.~\ref{fig:runtime_overview} draws that shared runtime from scrubbed
clinical input through a budgeted controller and module pool to one terminal
decision object.
Three inspectable principles organize the rest of this section.
\textbf{Shared Decision Object (SDO):} every registered surface returns one
decision object \(d\) under the same accept/abstain/escalate interface.
\textbf{Minimum-Sufficient Verification (MSV):} open only the modules needed under
budget \(b\), and stop once the decision is supported.
\textbf{Intermediate Repair (IRP):} repair may update the candidate inside the
loop, and never replaces the terminal decision object.
The subsections below formalize the shared objects under SDO, show a dermatology
case under IRP, then instantiate MSV sparse opening and IRP candidate updates
before listing the registered recipe adapters.

\subsection{Case record and static proposer}
\label{sec:method_overview}

Under SDO, clinical input in Fig.~\ref{fig:runtime_overview} enters as a
scrubbed case record and a static proposer emits the first candidate before any
module opens.
An incoming case is a source-identity-scrubbed case record
\begin{equation}
x=(v,q,o,m),
\end{equation}
where \(v\) is the image, \(q\) the request, \(o\) optional constraints, and
\(m\) deployment-visible metadata.
A static proposer \(P\) emits the initial candidate
\begin{equation}
a_0=P(x),
\end{equation}
where \(P\) is the proposer and \(a_0\) is its first-pass candidate.
Deterministic dispatch then selects a family--recipe route
\begin{equation}
r=(f,\rho)=R(x,a_0),
\end{equation}
where \(R\) is the dispatcher, \(f\) the family, and \(\rho\) the registered
recipe adapter.
Under budget \(b\), the runtime opens an ordered module history
\begin{equation}
H=(m_1,\ldots,m_K),
\end{equation}
where \(H\) is the opened-module history.
The unit module-call cost under the shared verification budget is then
\begin{equation}
C(H)=\sum_{k=1}^{K}c(m_k)\le b,
\end{equation}
where \(c(m)\) is the per-module charge with \(c(m)=1\) in primary comparisons.
Every case ends by locking one shared decision object that stores outcome fields
\begin{equation}
d=(y,s^{*},r,H,\tau),
\end{equation}
where \(y\) is the released or withheld answer, \(s^{*}\) is the terminal status, and \(\tau=\) the decision path tag.
That status ranges over the closed terminal set
\begin{equation}
s^{*}\in\{\mathrm{accept},\mathrm{abstain},\mathrm{escalate}\},
\end{equation}
where escalate covers out-of-profile or gate-failure cases as a non-accept reject
status.
Intermediate controls (continue / open-evidence / repair) may update the
candidate or history, but are never terminal outcomes.
Dermatology imaging on PAD-UFES-20 (PAD)~\cite{pacheco2020padufes} and IDRiD~\cite{porwal2018idrid} attach repair-first and post-repair specialist recipes.
Clinical image challenge (NEJM)~\cite{nejm_image_challenge} and HyperKvasir endoscopy (HK)~\cite{borgli2020hyperkvasir} attach knowledge-lock and conditional-specialist recipes under the same accept/repair/reject interface, so SDO holds across surfaces.
Algorithms~1--2 in \SuppSecappruntimeobjects{} expand the objects; runtime gate checks sit in \SuppSecappruntimegatechecklist{}.
That shared decision object keeps the released answer, the terminal status, and the opened history on every colored surface in that record.

\FloatBarrier
\subsection{Running example: dermatology repair}
\label{sec:method_running_example}

\begin{figure*}[t]
\setlength{\floatsep}{4pt plus0pt minus1pt}%
\setlength{\textfloatsep}{6pt plus0pt minus2pt}%
\setlength{\intextsep}{6pt plus0pt minus2pt}%
\centering
\includegraphics[width=\textwidth,keepaspectratio]{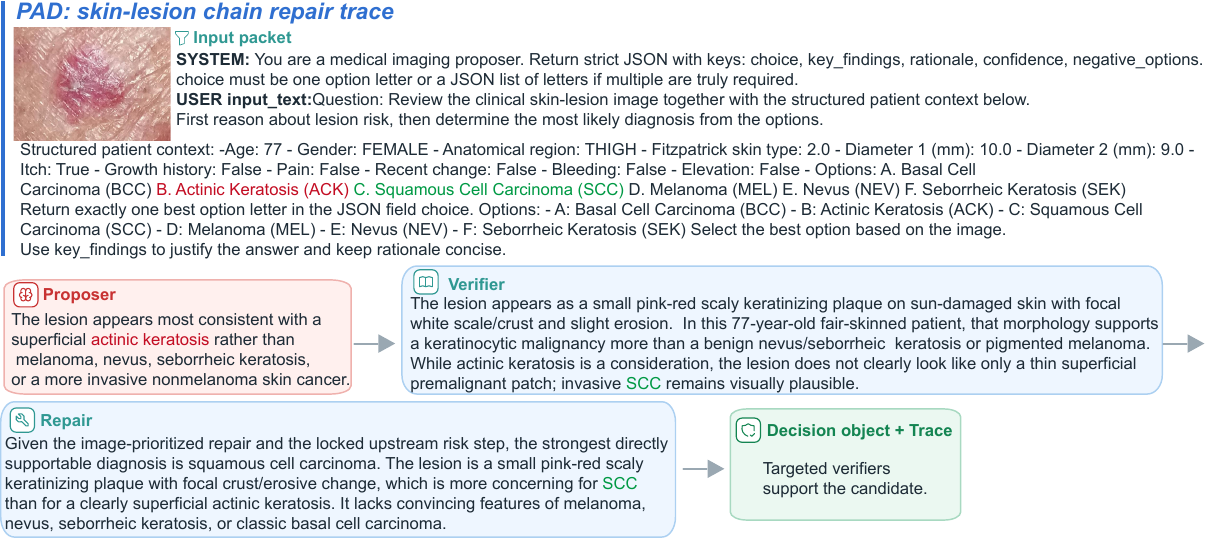}
\caption{PAD repair running example: wrong low clinical risk (low-risk) branch corrected
to squamous cell carcinoma, then accept with route and module history;
illustrative single-case trace.}
\label{fig:failure_mode_architectures}
\end{figure*}

The following running example is a single PAD~\cite{pacheco2020padufes} case under the registered dermatology recipe: IRP updates the candidate before the terminal SDO lock, matching Fig.~\ref{fig:failure_mode_architectures}.
A scrubbed record \(x\) hides source identity; proposer \(P\) emits a wrong low-risk candidate such as actinic keratosis, dispatch selects \(\rho_{\mathrm{derm}}\), and under budget \(b\) a failed cheap gate plus a verifier leads repair to write \(a_{t+1}\) as squamous cell carcinoma.
The lock then issues \(d=(y,s^{*},r,H,\tau)\) with accept, route, ordered history \(H\), and decision path tag \(\tau\), and \(C(H)\) charges each opened module once per call; if evidence stays insufficient the same lock abstains, as illustrated in Fig.~\ref{fig:failure_mode_architectures}.

\subsection{Budgeted sparse evidence cards}
\label{sec:method_shared_objects}

Under MSV, Fig.~\ref{fig:runtime_overview} prints seven shared-pool cards: Proposer emits
\(a_0\), Cheap gate can stop early, TextVerif checks text, Evidence opens
failure-mode evidence, Adjudicator resolves conflict, Repair updates
non-terminally, and Trace writes \(\tau\).
The same surface-colored dots appear on those cards and on the decision-object
Trace field, so which surfaces open each module and which surfaces emit a
decision path tag stay visible on the same cards.
Evidence/Repair trees open those cards by failure mode; Accept / Abstain /
Escalate are written by the decision lock as \(s^{*}\).
Modules expose comparable decision-facing signals so heterogeneous specialists
can share one fabric~\cite{lakshminarayanan2017deepensembles,guo2017calibration}.
Cheap misclassification scores stay gates rather than the terminal policy~\cite{hendrycks2017baseline,ovadia2019uncertainty}.
Knowledge evidence may reuse dense retrieval when heavier local modules stay closed~\cite{karpukhin2020dpr,borgeaud2022retro}.
Program-aided candidate updates stay inside the same repair loop~\cite{gao2023pal}.
After scrubbing, dispatch may use task content, modality, and answer schema but not source-dataset identity (\SuppSecappscrubbedrouting); the controller stays recipe- and threshold-based rather than end-to-end trained.
Strict / default / loose are \emph{evidence-budget} points on one accept policy and differ only in \(b\).
Given route \(r\), early accept, conflict escalate, repair-recheck, and budget-exhaust abstain under \(b\) follow the loop below; opened history is
\begin{equation}
H=\pi(r,x,a_0,b),
\end{equation}
where \(\pi\) is the shared control policy and \(x\) is the case record.
\par\noindent\fbox{%
\begin{minipage}{\dimexpr\linewidth-2\fboxsep-2\fboxrule\relax}\footnotesize
\textbf{Control loop under \(\pi\)}. Algorithms~1--2 in \SuppSecappruntimeobjects{} give the steps.
\begin{enumerate}[leftmargin=1.1em,itemsep=0pt,topsep=0pt,parsep=0pt,partopsep=0pt]
\item Scrub the case, then set the answer, route, recipe, and budget \(b\).
\item A cheap-gate pass keeps that answer and shuts heavier evidence.
\item Conflict opens verify, specialist or knowledge, then adjudicator.
\item A recoverable error updates the candidate and rechecks it.
\item Within budget \(b\), open required modules; otherwise lock abstain.
\item Both pools lock accept or abstain; repair is not terminal.
\end{enumerate}
\end{minipage}}
\par
Always-on (AO) enumeration versus sparse MedHarness (MH) opening under the shared policy is
\begin{equation}
H_{\mathrm{AO}}=\mathcal{M}_\rho,
\end{equation}
where \(H_{\mathrm{AO}}\) opens every recipe-admissible module under registered recipe adapter \(\rho\), and
\begin{equation}
\mbox{$H_{\mathrm{MH}}=\pi(r,x,a_0,b)$},
\end{equation}
where \(H_{\mathrm{MH}}\) is the sparse history under \(\pi\) and each opened module satisfies \(m_k\in\mathcal{M}_\rho\).
A known surface attaches by registration of its family and recipe
\begin{equation}
\Gamma(s)=(f,\rho),
\end{equation}
where \(\Gamma\) maps surface \(s\) to family--recipe \((f,\rho)\), so a newly
registered surface reuses the shared accept/repair/reject interface without a
private always-on stack.

\subsection{Intermediate repair under the same loop}
\label{sec:method_sparse_repair}

When opened evidence \(e_t\) supports a recoverable error, intermediate repair
updates the candidate inside the same loop before the terminal lock~\cite{cot,tot}:
\begin{equation}
a_{t+1}=\Psi(x,a_t,e_t),
\end{equation}
where \(\Psi\) is the intermediate candidate update; the lock assigns \(s^{*}\) from \((x,a_{t+1},e_t,H_{t+1})\) and repair never replaces \(d\) on that same case record.

\subsection{Instantiated recipes}
\label{sec:method_recipes}

The four recipes share the case record, router, and \(d\): dermatology repair-first,
retinal post-repair specialist, clinical question-answering (QA) cheap-first knowledge lock, and
endoscopy conditional specialist---adapters over one shared registration map,
not private always-on workflows.
A new surface attaches by \(\Gamma(s)\) rather than by rewriting that shared
decision object \(d\), so SDO remains the attachment rule.
\section{Experiments}

\subsection{Setup}
\label{sec:exp_setup}

We evaluate MedHarness as a budgeted runtime after a static proposer on medical
imaging surfaces: when to stop, open failure-mode evidence, repair, or reject
under one shared case-record interface.
Known registered recipes attach through that interface rather than discovering
surfaces zero-shot.
Comparisons ask three questions in order: whether one registered interface
attaches across heterogeneous decisions; whether budgeted sparse control
improves accepted quality versus always-on verification and selective
abstention on a shared proposer; and whether repair, specialist, and lock
levers account for that difference under the Method principles SDO, MSV, and IRP.
Those questions keep the same runtime reliability layer on every named pool.
Table~\ref{tab:eval_pools_and_abbrev} names the pools, surfaces, and comparison-arm Short|Meaning rows used below; correct accepts (CA) and wrong accepts (WA) follow \SuppSecappcoinedarmprotocols{}.
Table~\ref{tab:eval_pools_and_abbrev} sizes the All pool at 4848 from PAD~\cite{pacheco2020padufes} 200, IDRiD 259, NEJM 685, and HK 3704, and that table prints 384 on the Mix4 without HyperKvasir (Mix4-HKout) row, which is not the NEJM reweight proxy.
\SuppSecappadditionalresults{} names a reweight proxy whose NEJM count is not that surface size 685.
Pool composition and surface roles stay aligned with that table across every Results subsection that follows this setup.
Raw proposer floor (Prop.) denotes that comparison arm and always-open alias (Mono.) denotes that comparison arm in the \mbox{Short column below.}
Outcome counts (C/W/Abs.) record correct, wrong, and abstain.
Correct accepts (Correct/accepts) name that released-answer cell.
Self-consistency abstention (SC-Abs.) denotes that comparison arm.
Text-only verification (Text-check) denotes that comparison arm.
Agreement plus confidence (agree/conf) denotes that comparison arm.
Short evidence budget (early-stop) denotes that budget point.
Accepted accuracy (Acc.) and mean opened module calls (Mods.) are the metric
shorts used below.

\begin{table*}[t]
\centering
\caption{Evaluation pools and surfaces. Raw proposer floor (Prop.) and always open evidence alias (Mono.); Acc.; Mods.; always open evidence (Always-on); correct/wrong/abstain (C/W/Abs.); MH; rescue versus harm (R/H); Mix4 without HyperKvasir (Mix4-HKout); HyperKvasir holdout role (HK-out); HK; diabetic retinopathy (DR). Other arm and metric Short|Meaning pairs and open/stop detail follow \SuppSecappcoinedarmprotocols{}.}
\label{tab:eval_pools_and_abbrev}
\footnotesize
\renewcommand{\arraystretch}{0.90}
\setlength{\tabcolsep}{0.8pt}
\begin{tabularx}{\textwidth}{@{}lccc!{\color{black!28}\vrule width 0.35pt}lccc!{\color{black!28}\vrule width 0.35pt}l>{\raggedright\arraybackslash}X!{\color{black!28}\vrule width 0.35pt}l>{\raggedright\arraybackslash}X@{}}
\toprule
\multicolumn{4}{c}{\cellcolor{tablegray}\makebox[0.26\textwidth][c]{\textbf{\textit{Pools}}}} &
\multicolumn{4}{c}{\cellcolor{tablegray}\makebox[0.26\textwidth][c]{\textbf{\textit{Surfaces}}}} &
\multicolumn{4}{c}{\cellcolor{tablegray}\makebox[0.26\textwidth][c]{\textbf{\textit{Arms / Metrics}}}} \\
\cmidrule(lr){1-4}\cmidrule(lr){5-8}\cmidrule(lr){9-12}
\multicolumn{1}{c}{\cellcolor{tabgroupa}\textbf{Pool}} &
\multicolumn{1}{c}{\cellcolor{tabgroupa}\textbf{Role}} &
\multicolumn{1}{c}{\cellcolor{tabgroupa}\textbf{Comp}} &
\multicolumn{1}{c}{\cellcolor{tabgroupa}\textbf{$n$}} &
\multicolumn{1}{c}{\cellcolor{tabgroupb}\textbf{Surface}} &
\multicolumn{1}{c}{\cellcolor{tabgroupb}\textbf{Role}} &
\multicolumn{1}{c}{\cellcolor{tabgroupb}\textbf{Scope}} &
\multicolumn{1}{c}{\cellcolor{tabgroupb}\textbf{$n$}} &
\multicolumn{1}{c}{\cellcolor{tabgroupa}\textbf{Short}} &
\multicolumn{1}{c}{\cellcolor{tabgroupa}\textbf{Meaning}} &
\multicolumn{1}{c}{\cellcolor{tabgroupb}\textbf{Short}} &
\multicolumn{1}{c}{\cellcolor{tabgroupb}\textbf{Meaning}} \\
\midrule
All & Attach+replay & 4 surf & 4848 &
PAD & Skin repair & skin & 200 &
Prop. & Raw proposer floor &
Acc. & Correct/accepts \\
Live512 & Stop; HK out & 171/171/170 & 512 &
IDRiD & Retinal & 5 DR & 259 &
Mono. & Always open alias &
Mods. & Module charge \\
Mix4 & Stress+HK & 128$\times$4 & 512 &
NEJM & Quiz+know. & quiz & 685 &
Always-on & Full evidence open &
C/W/Abs. & Correct/wrong/abstain \\
Mix4-HKout & HK-out & 128$\times$3 & 384 &
HK & Endoscopy & endo & 3704 &
MH & Sparse controller &
R/H & Rescue versus harm \\
\bottomrule
\end{tabularx}
\end{table*}

\paragraph{Pools and surfaces.}
Table~\ref{tab:eval_pools_and_abbrev} defines Live512 as the primary quality pool for when-to-stop decisions, with 171/171/170
PAD~\cite{pacheco2020padufes}/IDRiD/NEJM cases and HyperKvasir held out.
The same table defines Mix4 as the complementary stress pool with 128 cases per family and HyperKvasir
included; Mix4-HKout holds HyperKvasir out for specialist-held stress.
PAD--UFES-20 follows prior skin-surface releases~\cite{pacheco2020padufes}.
Companion dermatology imaging context follows~\cite{tschandl2018ham10000}.
IDRiD follows the retinal grading benchmark~\cite{porwal2018idrid,ting2017dr}.
HyperKvasir follows the endoscopy benchmark~\cite{borgli2020hyperkvasir,pogorelov2017kvasir}.
NEJM Image Challenge follows the public clinical image quiz~\cite{nejm_image_challenge}. Those public datasets supply the imaging surfaces; Live512 and Mix4 are evaluation pools and task splits we organize for quality versus stress comparisons.
Reviews of language models in medicine give extra context for agent stacks~\cite{thirunavukarasu2023llms}.
Cases enter as source-identity-scrubbed case records; the proposer, scorer,
module pool, and runtime stay unchanged unless a row states otherwise.
Primary tables use one strong proprietary proposer GPT-5.4.
On the same Live512 ids, a Grok-4.5 stack transfer keeps specialists unchanged
while swapping the LLM proposer, verifier, repair, and lock stack
(\SuppSecappalternateprovidertransfer{}).
Every main comparison holds the proposer unchanged and changes only which evidence
path the registered controller may open under the shared verification budget on each scored arm.

\paragraph{Proposer stack and temperatures.}
Unless noted, the proposer is GPT-5.4; verification, repair, and lock
use the same stack.
Proposer temperature is \(0.1\); verifier, repair, and lock temperature is \(0.0\).
Specialists and role prompts sit in \SuppSecappcomponentcontracts.
Recipes and thresholds were set offline before Mix4 and Live512, then held
unchanged for every comparison arm in the Results tables below.
No main comparison changes the proposer mid-table.

\subsection{Evaluation metrics and accounting}
\label{sec:exp_protocol}

Main comparisons match proposer-only against full MedHarness on each named pool
under the same accounting.
Arm and metric shorts follow \SuppSecappcoinedarmprotocols{}; the
full calculus expands in \SuppSecappdecisioncalculus{}.
Acc. is correct accepts among accepts; Cov.\ and risk among accepts (Rsk.) appear where dual accept-share and risk columns print Cov/Rsk.
Correct/wrong/abstain counts (C/W/Abs.) record correct accepts, wrong accepts, and abstentions; Mods. is the mean opened-module charge; R/H and R-H score rescue versus harm against the proposer floor.
The three named pools stay non-interchangeable under the Setup pool composition
above: the Live512 quality pool and the Mix4 stress ranking share the accuracy column and the module column in the result tables.
Selective matching, bootstrap intervals, and repair/specialist/lock ablations
isolate levers; leave-one-family-out, five-split, and reweight rows stay supplementary
sensitivity (\SuppSecappadditionalresults{}).
Pool roles stay distinct in \SuppSecappwhylivenemix{}.
Unless noted, every comparison arm uses the same held recipe thresholds and
proposer--scorer pair on its named pool, with no per-arm retuning of decoding or
specialist role prompts on that pool for every comparison arm.

\section{Results}
Results climb from multi-surface decision quality on Live512, through same-pool
activation policy that isolates MSV, Mix4 stress efficiency under HyperKvasir,
lever ablation that answers SDO/MSV/IRP, and default budget operating point checks
that locate the evidence-budget setting on Mix4 and Live512.

\subsection{Multi-surface decision quality}
\label{sec:results_surfaces}

\textbf{Live512 quality.}\label{sec:results_surfaces_pool_quality} Primary when-to-stop quality on the held-out pool establishes the shared controller win against always-on verification.
We show that abstention on Live512, with HyperKvasir held out, cuts wrong accepts from 139 to 74. Correct accepts move from 373 to 367 with 71 abstentions, while Acc.\ is 0.832 at Cov.\ 0.861 (367/74/71) versus full verification Acc.\ 0.729 at Cov.\ 1.00 (373/139/0), and module means are 3.86 versus 4.00 (Table~\ref{tab:main_results}).
Repair keeps wrong accepts at 74 while moving the No repair row from 312/74/126 at 0.808 to 367 correct accepts and 71 abstentions at 0.832 (\SuppSecappadditionalresults{}).
Risk among accepts falls from 0.271 to 0.168 on Live512 at Cov.\ 0.861 versus the full-verification baseline at Cov.\ \mbox{1.00 (Table~\ref{tab:main_results}).}

\begin{table*}[t]
\centering
{\footnotesize
\caption{Live512 quality and Mix4 stress versus full verification and selective arms. Surface rates under 171/171/170 match the pool. HK\textdaggerdbl\ stays outside that stop denominator. Correct/wrong/abstain (C/W/Abs.); correct (C), wrong (W), rescue (R), harm (H); mean opened module calls (Mods.); rescue versus harm (R/H); rescue minus harm (R-H); proposer only floor (Proposer-only); always open evidence arm (Always-on); self consistency abstention (SC-Abs.); text only verification (Text-check); MedHarness minus proposer (MH-Prop.); MedHarness minus always on (MH-Mono.). Arm procedures follow \SuppSecappcoinedarmprotocols{}.}
\label{tab:main_results}
\renewcommand{\arraystretch}{0.90}
\setlength{\tabcolsep}{0.8pt}
\begin{tabularx}{\textwidth}{@{}l*{4}{>{\centering\arraybackslash}X}!{\color{black!28}\vrule width 0.35pt}*{3}{>{\centering\arraybackslash}X}@{}}
\toprule
& \multicolumn{4}{c}{\cellcolor{tablegray}\makebox[0.47\textwidth][c]{\textbf{\textit{Live512 surface C\%/W\%}}}} &
\multicolumn{3}{c}{\cellcolor{tablegray}\makebox[0.35\textwidth][c]{\textbf{\textit{Live512 Pool}}}} \\
\cmidrule(lr){2-5}\cmidrule(lr){6-8}
\multicolumn{1}{c}{\normalsize Arm} &
% table-header seed cites (non-printing): PAD\cite{pacheco2020padufes}; IDRiD\cite{porwal2018idrid}; NEJM\cite{nejm_image_challenge}
\multicolumn{1}{c}{\cellcolor{tabgroupa}\footnotesize PAD C\%/W\%} &
\multicolumn{1}{c}{\cellcolor{tabgroupb}\footnotesize IDRiD C\%/W\%} &
\multicolumn{1}{c}{\cellcolor{tabgroupa}\footnotesize NEJM C\%/W\%} &
\multicolumn{1}{c}{\cellcolor{tabgroupb}\footnotesize HK\textdaggerdbl\ C\%/W\%} &
\multicolumn{1}{c}{\cellcolor{tabgroupa}\footnotesize C/W/Abs.} &
\multicolumn{1}{c}{\cellcolor{tabgroupa}\footnotesize Acc.$\uparrow$} &
\multicolumn{1}{c}{\cellcolor{tabgroupa}\footnotesize Mods.$\downarrow$} \\
\midrule
Proposer-only & 76.6/23.4 & 35.7/64.3 & 64.1/35.9 & 43.9/56.1 & 301/211/0 & 0.588 & 0.00 \\
Always-on & 85.4/14.6 & 50.3/49.7 & 82.9/17.1 & 42.2/57.8 & 373/139/0 & 0.729 & 4.00 \\
\rowcolor{tableblue}
\textbf{MedHarness} & \textbf{83.6/12.3} & \textbf{50.3/12.3} & \textbf{81.2/18.8} & \textbf{93.6/0.0} & \textbf{367/74/71} & \textbf{0.832}\ifmhCompareDeltas\mhDeltaImproveUp{+0.103}\fi & \textbf{3.86}\ifmhCompareDeltas\mhDeltaImproveDown{-0.14}\fi \\
\bottomrule
\end{tabularx}%
\par\nointerlineskip
\setlength{\tabcolsep}{0.8pt}%
{\mhCompareDeltasfalse
\begin{tabularx}{\textwidth}{@{}l>{\centering\arraybackslash\hsize=1.20\hsize}X*{3}{>{\centering\arraybackslash\hsize=0.93333\hsize}X}!{\color{black!28}\vrule width 0.35pt}l>{\centering\arraybackslash\hsize=1.20\hsize}X*{3}{>{\centering\arraybackslash\hsize=0.93333\hsize}X}!{\color{black!28}\vrule width 0.35pt}l*{2}{>{\centering\arraybackslash}X}@{}}
\multicolumn{5}{c}{\cellcolor{tablegray}\makebox[0.36\textwidth][c]{\textbf{\textit{Acc./Cov./Risk}}}} &
\multicolumn{5}{c}{\cellcolor{tablegray}\makebox[0.36\textwidth][c]{\textbf{\textit{Mix4 Ranking}}}} &
\multicolumn{3}{c}{\cellcolor{tablegray}\makebox[0.20\textwidth][c]{\textbf{\textit{Bootstrap}}}} \\
\cmidrule(lr){1-5}\cmidrule(lr){6-10}\cmidrule(lr){11-13}
\multicolumn{1}{c}{\cellcolor{tabgroupb}\footnotesize Arm} &
\multicolumn{1}{c}{\cellcolor{tabgroupb}\footnotesize C/W/Abs.} &
\multicolumn{1}{c}{\cellcolor{tabgroupb}\footnotesize Acc.$\uparrow$} &
\multicolumn{1}{c}{\cellcolor{tabgroupb}\footnotesize Cov} &
\multicolumn{1}{c}{\cellcolor{tabgroupb}\footnotesize Risk$\downarrow$} &
\multicolumn{1}{c}{\cellcolor{tabgroupa}\footnotesize Arm} &
\multicolumn{1}{c}{\cellcolor{tabgroupa}\footnotesize C/W/Abs.} &
\multicolumn{1}{c}{\cellcolor{tabgroupa}\footnotesize Acc.$\uparrow$} &
\multicolumn{1}{c}{\cellcolor{tabgroupa}\footnotesize R/H} &
\multicolumn{1}{c}{\cellcolor{tabgroupa}\footnotesize Mods.$\downarrow$} &
\multicolumn{1}{c}{\cellcolor{tabgroupb}\footnotesize Contrast} &
\multicolumn{1}{c}{\cellcolor{tabgroupb}\footnotesize $\Delta$} &
\multicolumn{1}{c}{\cellcolor{tabgroupb}\footnotesize CI} \\
\midrule
Always-on & 373/139/0 & 0.729 & 1.00 & 0.271 &
Proposer-only & 256/256/0 & 0.500 & 0/0 & 0.00 &
MH-Prop.\ CA & +154 & 132/177 \\
\cellcolor{tableblue}\textbf{MedHarness} & \cellcolor{tableblue}\textbf{367/74/71} & \cellcolor{tableblue}\textbf{0.832} & \cellcolor{tableblue}\textbf{0.861} & \cellcolor{tableblue}\textbf{0.168} &
SC-Abs. & 251/194/67 & 0.564 & 35/15 & 0.00 &
MH-Prop.\ WA & $-$210 & -231/-188 \\
SC-Abs. & 258/209/45 & 0.552 & 0.912 & 0.448 &
Text-check & 261/245/6 & 0.516 & 37/31 & 1.99 &
MH-Prop.\ R-H & +166 & 145/187 \\
Text-check & 277/234/1 & 0.542 & 0.998 & 0.458 &
Always-on & 307/205/0 & 0.600 & 76/25 & 4.00 &
MH-Mono.\ CA & +103 & 82/124 \\
& & & & &
\cellcolor{tableblue}\textbf{MedHarness} & \cellcolor{tableblue}\textbf{410/46/56} & \cellcolor{tableblue}\textbf{0.899} & \cellcolor{tableblue}\textbf{168/2} & \cellcolor{tableblue}\textbf{3.15} &
MH-Mono.\ WA & $-$159 & -181/-139 \\
\bottomrule
\end{tabularx}
}% mhCompareDeltasfalse
}
\end{table*}

\textbf{Surface attachments.}\label{sec:results_surfaces_attachments} PAD~\cite{pacheco2020padufes}, IDRiD, NEJM, and HyperKvasir attach four surfaces of the same runtime decision under SDO rather than separate specialty stacks.
We show that those four attachments improve one shared decision interface. The pooled Live512 result remains 367/74/71.
PAD~\cite{pacheco2020padufes} recovers 83.6/12.3 versus the proposer 76.6/23.4, beside always-on 85.4/14.6 on Table~\ref{tab:main_results}.
IDRiD keeps correct-accept 50.3 and cuts wrong accepts from always-on 49.7 to 12.3 on Table~\ref{tab:main_results}.
On IDRiD Acc.\ among accepts rises to 80.4 from proposer 35.7; pool correct-accept rises to 50.3 versus 35.7, so the quality gain combines more correct accepts with explicit abstention at 37.4 Abs.\ that cuts wrong accepts from 64.3 to 12.3.
NEJM reaches 81.2/18.8 versus the proposer 64.1/35.9, beside always-on 82.9/17.1 on Table~\ref{tab:main_results}.
The registration-scale marker on Table~\ref{tab:main_results} prints 93.6/0.0 and 43.9/56.1 beside always-on 42.2/57.8, and that marker is the one defined in \SuppSecappcoinedarmprotocols{}.
Those rates are separate from the 128-case rows 119/0/9 and 64/64/0.
Thus absolute surface rates are heterogeneous; the shared metric is decision quality under a static proposer on each registered surface in clinical decision workflows~\cite{tu2025amie}.

\textbf{Selective foils.}\label{sec:results_surfaces_foils} Printed selective rows on Live512 isolate the gap below the proposer. We show that MedHarness beats that non-abstaining foil on the matched wrong-accept row.
On Table~\ref{tab:main_results}, Live512 self-consistency abstention reaches Acc.\ 0.552 and the text check gate reaches Acc.\ 0.542, both under the proposer Acc.\ 0.588, and matching the MedHarness wrong-accept count collapses both arms so they abstain on every case.
Thus the matched wrong-accept foil is the printed row in \SuppSecappadditionalresults{}.
This means the full 4848-case scrubbed replay is path-complete under registered routing with preserved surface formats (\SuppSecappscrubbedrouting{}).
Case panels and style galleries under the same controller preserve the accept/repair/abstain story in \SuppSecappcasestudies{}, with the Live512 quality \mbox{rows on Table~\ref{tab:main_results}.}

\subsection{Same pool activation policy}
\label{sec:results_activation}

\begin{figure*}[t]
\setlength{\floatsep}{4pt plus0pt minus1pt}%
\setlength{\textfloatsep}{6pt plus0pt minus2pt}%
\setlength{\intextsep}{6pt plus0pt minus2pt}%
\centering
\includegraphics[width=\textwidth,keepaspectratio]{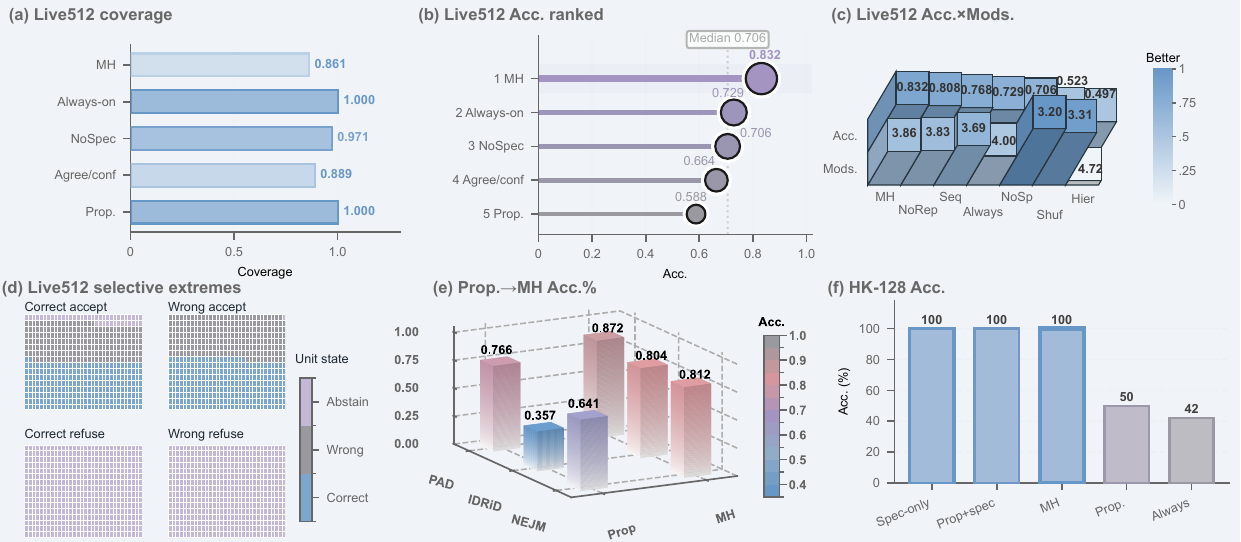}
\caption{Live512 support overview; HyperKvasir (HK) sits outside the primary stop pool: (a) Live512 Cov.; (b) Live512 Acc.\ ranked; (c) Live512 Acc.$\times$Mods.\ lever heatmap; (d) Live512 selective extremes; (e) Prop.\ to MedHarness accepted Acc.\ by surface; (f) HK 128 Acc.\ by comparison arms. Panel ink: proposer only floor (Prop.), agreement plus confidence (agree/conf), always-open evidence arm (Always-on), No specialist control (NoSpec), MedHarness (MH), correct/wrong/abstain counts (C/W/Abs.), mean opened module calls (Mods.); specialist only, sequence lock (Seq), shuffled route (Shuf), hierarchical adapted (Hier), and proposer plus specialist open/stop detail \SuppSecappcoinedarmprotocols. Complements Table~\ref{tab:main_results}.}
\label{fig:live512_support6}
\end{figure*}

\textbf{Sparse activation.}\label{sec:results_activation_sparse_win} An unchanged module pool locks the Live512 wrong-accept win so MSV cannot be explained by opening extra branches.
We show that sparse conditional activation on Live512 cuts wrong accepts versus always-on verification under the shared module pool.
MedHarness reaches 367/74/71 against always-on 373/139/0 on Table~\ref{tab:main_results}. Agreement plus confidence (agree/conf) on Fig.~\ref{fig:live512_support6} fails to match that wrong-accept point.
\textbf{Activation panels.}\label{sec:results_activation_fig3} Cov., ranked Acc., lever heatmap, and accepted Acc.\ by surface establish the same sparse win visually without inventing a second pool.
We show that Fig.~\ref{fig:live512_support6}(a) improves the comparison. Cov.\ on Fig.~\ref{fig:live512_support6}(a) stays below full verification on the same Live512 cases, and Fig.~\ref{fig:live512_support6}(b) keeps the accepted-answer rank.
Lever and selective panels keep specialist cells on Fig.~\ref{fig:live512_support6}(c) and extremes on Fig.~\ref{fig:live512_support6}(d).
Surface and HK panels keep accepted Acc.\ by surface on Fig.~\ref{fig:live512_support6}(e) and HK 128 Acc.\ on Fig.~\ref{fig:live512_support6}(f) for specialist only, proposer plus specialist, and MedHarness.
The 128-case controls are specialist only 128/0/0, proposer plus specialist 121/0/7, MedHarness 119/0/9, proposer only 64/64/0, and always-on 54/74/0.
Fig.~\ref{fig:live512_support6}(f) prints 100 for specialist-only, proposer-plus-specialist, and MedHarness, prints 50 for the proposer, and prints 42 for always-on, and the specialist-only count row already prints zero wrong accepts.
Thus sparse conditional activation under a shared module pool beats always-on opening and agreement plus confidence cascades on the same Live512 cases, answering MSV without changing the proposer or the admissible module pool.
This means activation policy alone, under MSV, moves the quality point when the module pool is held unchanged.
This means Mix4-HKout and no-specialist activation controls under the same shared module pool stay in \SuppSecappuactivationpolicysamepool{}.
\subsection{Mix4 stress efficiency}
\label{sec:controlled_activation_value}

\begin{figure*}[t]
\setlength{\floatsep}{4pt plus0pt minus1pt}%
\setlength{\textfloatsep}{6pt plus0pt minus2pt}%
\setlength{\intextsep}{6pt plus0pt minus2pt}%
\centering
\includegraphics[width=\textwidth,keepaspectratio]{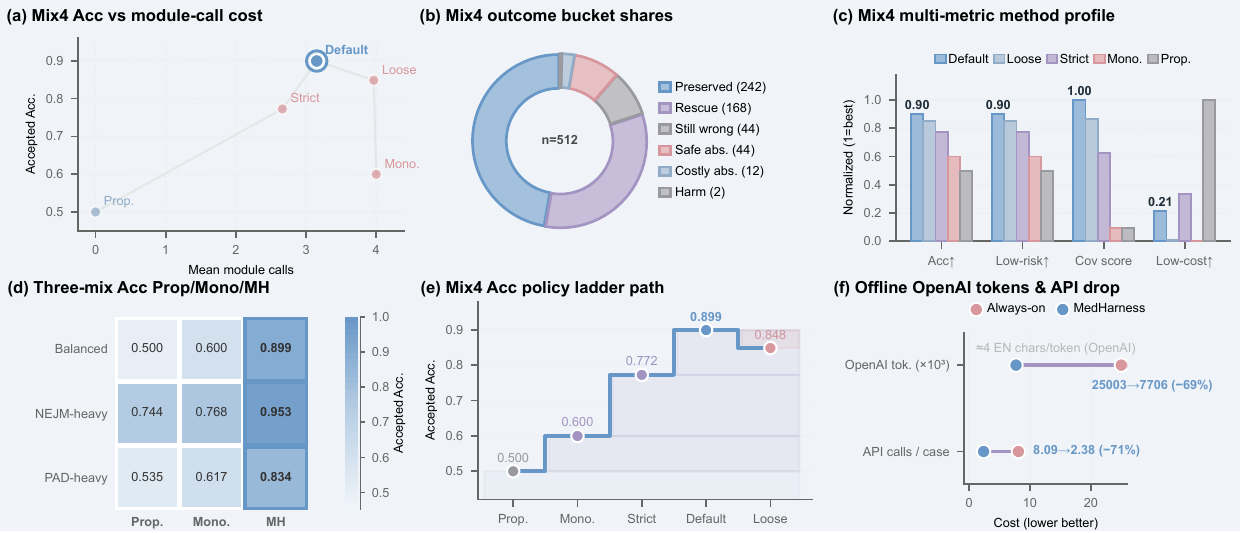}
\caption{Mix4 stress ranking; HyperKvasir (HK) is included: (a) Acc.\ vs module call cost; (b) outcome bucket shares; (c) multi metric method profile; (d) Acc.\ across three Mix4 mixes for Prop./Mono./MedHarness (MH); (e) Acc.\ policy ladder; (f) offline tokens and API call drop under OpenAI tokenizer guidance (OpenAI). Panel ink: proposer only floor (Prop.), Always-on alias (Mono.), always-open evidence arm (Always-on), MedHarness (MH); open/stop detail \SuppSecappcoinedarmprotocols{}. Complements Table~\ref{tab:main_results} confidence intervals.}
\label{fig:mix4_merge_support6}
\end{figure*}

\textbf{Mix4 stress ranking.}\label{sec:results_mix4_ranking} Failure-enriched family balance isolates efficiency so the sparse Default point cannot be read as a restatement of the Live512 quality pool.
We show that budgeted MedHarness beats always-on and the proposer floor on Mix4 Acc./Mods.\ under HyperKvasir-inclusive stress (Table~\ref{tab:main_results}).
The pool holds four families with 128 cases each, and the 50/50 balance and the rescue counts are against the cached proposer floor of sixty-four proposer-correct and sixty-four proposer-wrong per family, hence the cached 256/256/0 activation floor, beside a separate live GPT-5.4 remeasure 274/238/0 (Acc.\ 0.535) on the same ids, and the paired MH-Prop.\ deltas and R/H 168/2 stay on that cached proposer floor.
Fig.~\ref{fig:mix4_merge_support6}(b) prints Rescue (168) and Harm (2).
Proposer-only is 256/256/0 at 0 modules; the always-open evidence arm is 307/205/0; MedHarness is 410/46/56.
Mix4's module column includes the HyperKvasir specialist used for this stress ranking; Live512 quality remains on Table~\ref{tab:main_results}.
\textbf{Offline cost.}\label{sec:results_mix4_offline_cost} Token and API call cuts under matched context-load accounting travel with the same Default budget rather than a separate cheap-only recipe.
We show that MedHarness cuts those totals. Calls go from 8.09 to 2.38 (~71\%) and offline tokens from 25003 to 7706 (~69\%) on Fig.~\ref{fig:mix4_merge_support6}(f) under matched context-load accounting that estimates tokens at about 4 English characters per token following OpenAI tokenizer guidance, distinct from Live512 provider billing.
The five bootstrap intervals in Table~\ref{tab:main_results} pair to the printed Mix4 contrasts on that table: the MH-Prop.\ rows, the rescue count, and R/H 168/2 use the cached proposer floor 256/256/0, with MH-Prop.\ deltas of +154 correct accepts, minus 210 wrong accepts, and +166 rescue-minus-harm.

\textbf{Default operating point.}\label{sec:results_mix4_fig4_bind} Module cost, outcome shares, and mix slices establish the Default sparse point so later ablation and budget subsections stress one shared operating point.
We show that default sparse control on Mix4 with HyperKvasir beats always-on. Acc.\ is 0.899 with 46 wrong accepts against always-on Acc.\ 0.600 with 205 wrong accepts on Table~\ref{tab:main_results}.
Cost and share panels keep the Default point on Fig.~\ref{fig:mix4_merge_support6}(a) Acc.\ versus module-call plane and Fig.~\ref{fig:mix4_merge_support6}(b) outcome shares.
Profile and mix panels keep multi-metric ranking on Fig.~\ref{fig:mix4_merge_support6}(c) and three-mix Acc.\ slices on Fig.~\ref{fig:mix4_merge_support6}(d).
Fig.~\ref{fig:mix4_merge_support6}(d) prints the raw proposer, Prop., on that three-mix panel, a different quantity from the live remeasure count already stated for the unpaired ids.
Ladder and offline panels keep the Acc.\ rank on Fig.~\ref{fig:mix4_merge_support6}(e) and the token and API drop under OpenAI tokenizer guidance on Fig.~\ref{fig:mix4_merge_support6}(f), while leave-one-family and five-split rechecks stay in \SuppSecsecsuppmixbudgetladder{} under that same Default ladder. Thus that offline Mix4 estimate stays on the offline panel, as printed with those Mix4 call and token cuts on the offline panel.
\subsection{Lever ablation}
\label{sec:results_ablation}

\textbf{Principle levers.}\label{sec:results_ablation_levers} Gold recipe binding establishes the full sparse Mix4 point on the printed main-table row. We show that lever removals in this ablation cut printed Live512 rows.
No specialist is 351/146/15, no repair is 312/74/126, sequence lock off is 393/119/0, and shuffled route is 196/179/137 on the Live512 rows, beside MedHarness 367/74/71 and always-on 373/139/0.
The always-open evidence arm (Always-on) is 307/205/0 on this stress pool on Table~\ref{tab:main_results}, and MedHarness reaches the Mix4 sparse point at 410/46/56 on that table.
Mix4 shuffled binding stays off Table~\ref{tab:main_results}, and route-only sparse stays off that table.
Thus route control under SDO attachment sets that sparse Mix4 point~\cite{zhao2026deeprare}; hierarchy Acc. ladders in \SuppSecappadditionalresults{} and the capability contrast in \SuppSecappaboveprocapability{} \mbox{keep the same controller ranking.}

\subsection{Default budget operating point}
\label{sec:results_budget}

\textbf{Default utility.}\label{sec:results_budget_utility} The Default evidence budget locks the Mix4 utility peak that later leave-out and transfer checks must keep ranking under the same controller.
We show that the MedHarness default budget improves the peak. Counts stay 410/46/56 on the Mix4 ladder, and Utility 174.6 is the author-chosen weighted score of that same budget in \SuppSecsecsuppmixbudgetladder{}; the weights sit in \SuppSecappdecisioncalculus{}.
Thus default Mix4 evidence budgeting raises stress-pool utility over early-stop and extended-evidence regimes, while Live512 keeps the when-to-stop quality win under the same accept/repair/reject interface.
Offline API/token cuts under that default (~71\% calls, ~69\% tokens) stay in Fig.~\ref{fig:mix4_merge_support6}(f).

\textbf{Held-out checks.}\label{sec:results_budget_held} Family leave-out and Mix4-without-HyperKvasir establish the Default ranking beyond a single pool. The alternate language-model transfer beats always-on verification and the proposer.
The alternate-model transfer prints MedHarness 306/113/93 at Acc.\ 0.730, the always-open evidence arm 341/171/0 at Acc.\ 0.666 and Proposer-only 303/209/0 at Acc.\ 0.592; clinician reviews likewise stress multi-threshold operating points~\cite{liu2019comparison}. Those rankings name MedHarness 306/113/93, the always-open evidence arm 341/171/0, and Proposer-only 303/209/0.
\section{Conclusion}
\label{sec:discussion}
In this work, we introduce MedHarness, a budgeted multi-surface runtime that reformulates reliable medical imaging agent control as registered surface attachment under a shared policy with sparse verification and \mbox{those accept/repair/reject decisions.}
Guided by SDO, MSV, and IRP, MedHarness reaches Acc.\ 0.832 on Live512 and Acc.\ 0.899 on Mix4 under the same controller while cutting wrong accepts and remote-service load versus always-open full verification on the registered pools (Table~\ref{tab:main_results}).
Live ablations isolate repair, specialist evidence, and route binding as the levers that set that sparse operating point across the shared module fabric.

Limitations include that unseen surfaces still require a registered recipe adapter under shared \(\pi\), and that assessing prospective clinical use without live billing evidence remains open (\SuppSecappclaimscopeparked{}, \SuppSecappreproducibilityboundary{}; \cite{liu2020consortai,vasey2022decideai}).
Prospective clinical-trial guidance for AI interventions remains relevant here~\cite{rivera2020spiritai,nagendran2020diagnostic}.
Present validation focuses on offline registered pools, and further evaluation can cover surfaces beyond these pools.
This preprint contains the main text only. The supplementary sections cited above are not included in this PDF.

\bibliographystyle{iclr2027_conference}
\bibliography{main}
\end{document}